\documentclass[lettersize,journal]{IEEEtran}
\usepackage{amsmath,amsfonts}
\usepackage{algorithmic}
\usepackage{algorithm}
\usepackage{array}
\usepackage[caption=false,font=normalsize,labelfont=sf,textfont=sf]{subfig}
\usepackage{textcomp}
\usepackage{stfloats}
\usepackage{url}
\usepackage{verbatim}
\usepackage{graphicx}
\usepackage{cite}
\usepackage{amssymb}
\usepackage{amsmath}
\usepackage{amsthm}
\newtheorem{theorem}{Theorem}

\usepackage{algorithm}
\usepackage{algorithmic}
\usepackage{booktabs}
\begin{document}

\title{An Ordinary Differential Equation Sampler with Stochastic Start for Diffusion Bridge Models}

\author{Yuang Wang, Pengfei Jin, Li Zhang, Quanzheng Li, Zhiqiang Chen and Dufan Wu
\thanks{Send correspondence to Dufan Wu (E-mail: dwu6@mgh.harvard.edu).

Yuang Wang, Li Zhang and Zhiqiang Chen are with the Department of Engineering Physics, Tsinghua University, Beijing, 100084, China. 
Yuang Wang, Pengfei Jin, Quanzheng Li and Dufan Wu are (were) with the Department of Radiology, Massachusetts General Hospital and Harvard Medical School, Boston, MA 02114, USA.
}}
\markboth{ODE Sampler with Stochastic Start for Diffusion Bridge Models}%
{An Ordinary Differential Equation Sampler with Stochastic Start for Diffusion Bridge Models}


\maketitle

\begin{abstract}
Diffusion bridge models have demonstrated promising performance in conditional image generation tasks, such as image restoration and translation, by initializing the generative process from corrupted images instead of pure Gaussian noise. However, existing diffusion bridge models often rely on Stochastic Differential Equation (SDE) samplers, which result in slower inference speed compared to diffusion models that employ high-order Ordinary Differential Equation (ODE) solvers for acceleration. To
mitigate this gap, we propose a high-order ODE sampler with a stochastic start for diffusion bridge models. To overcome the singular behavior
of the probability flow ODE (PF-ODE) at the beginning of the
reverse process, a posterior sampling approach was introduced at the first reverse step. The sampling was designed to ensure a smooth transition from corrupted images to the
generative trajectory while reducing discretization errors. Following this stochastic start, Heun's second-order solver is applied to solve the PF-ODE, achieving high perceptual quality with significantly reduced neural function evaluations (NFEs). Our method is fully compatible with pretrained diffusion bridge models and requires no additional training. Extensive experiments on image restoration and translation tasks, including super-resolution, JPEG restoration, Edges-to-Handbags, and DIODE-Outdoor, demonstrated that our sampler outperforms state-of-the-art methods in both visual quality and Frechet Inception Distance (FID).
\end{abstract}


\section{Introduction}
\IEEEPARstart{S}{CORE}-based diffusion models\cite{song2020score,dhariwal2021diffusion,karras2022elucidating, karras2024analyzing}, grounded in stochastic theory, map Gaussian noise to the data distribution via learned score functions, and have achieved the state-of-the-art performance in image generation tasks. Compared to Generative Adversarial Networks (GANs)\cite{wang2021real,isola2017image}, diffusion models provide superior perceptual quality and greater stability when sampling from complex data distributions. However, initializing the generative process from pure Gaussian noise can be suboptimal for conditional generation tasks like image restoration and translation, as corrupted images contain significantly more structural information than random noise. Diffusion bridge models address this limitation by starting the generative process from corrupted images, leveraging their structural similarity to clean images. Recent advances, such as the Image-to-Image Schrödinger Bridge (I$^2$SB)\cite{liu20232} and the Denoising Diffusion Bridge Model (DDBM)\cite{zhou2023denoising}, have demonstrated substantial performance improvements over traditional diffusion models in these applications.

Despite these advancements, existing diffusion bridge models, including Inversion by Direct Iteration (InDI) \cite{delbracio2023inversion}, I$^2$SB, and Consistent Direct Diffusion Bridge (CDDB) \cite{chung2024direct}, predominantly rely on Stochastic Differential Equation (SDE) samplers. This reliance results in slower inference speeds compared to diffusion models that adopt high-order Probability Flow Ordinary Differential Equation (PF-ODE) solvers for acceleration\cite{karras2022elucidating,lu2022dpm,liu2022pseudo}. Notably, while DDBM highlights that pure ODE samplers often produce blurry images in diffusion bridge models, it did not identify the underlying cause. To address this issue, DDBM employs a high-order hybrid strategy alternating between ODE and SDE samplers. It still requires over 100 neural function evaluations (NFEs) to achieve satisfactory results.

In this work, we identify that the limited performance of pure ODE samplers in diffusion bridge models arises from the singular behavior of the PF-ODE at the start of the generative process. Leveraging this insight, we propose an ODE Sampler with a Stochastic Start (ODES3) for diffusion bridge models. Our approach employs posterior sampling to transition corrupted images into intermediate representations where the PF-ODE becomes well-defined. Subsequently, we apply the second-order Heun solver\cite{ascher1998computer} to the PF-ODE, achieving high perceptual quality with reduced NFEs. Our sampler is fully compatible with pretrained diffusion bridge models and requires no additional training. Its effectiveness is validated on image restoration and translation tasks using pretrained models from I$^2$SB and DDBM. The proposed sampler outperforms the original samplers used in I$^2$SB and DDBM, as well as other state-of-the-art methods, in terms of both Frechet Inception Distance (FID)\cite{heusel2017gans} and visual quality.

\section{Related Work}
In this section, we review key acceleration strategies in diffusion models and recent advancements in diffusion bridge models.
\subsection{Acceleration for Diffusion Models}

Accelerating the inference process of diffusion models has become a critical area of research. DDIM\cite{song2020denoising} was the first to address this challenge by transforming the generative process of DDPM\cite{ho2020denoising} from a Markovian to a non-Markovian framework, establishing its connection to the PF-ODE. ScoreSDE\cite{song2020score} further bridged DDPM and score-based models, showing that the reverse process can be described using either the reverse SDE or the PF-ODE. Acceleration strategies based on the PF-ODE can be broadly classified into two categories.

The first category involves methods that require no additional training and leverage high-order ODE solvers to speed up the generative process. For example, EDM\cite{karras2022elucidating} and EDM2\cite{karras2024analyzing} utilize Heun’s second-order method to solve the PF-ODE, achieving state-of-the-art performance in image generation. Additionally, approaches such as DPM-solver\cite{lu2022dpm} and PNDM\cite{liu2022pseudo} employ exponential integrators to reduce discretization errors, ensuring that intermediate images remain consistent with the true noisy manifold.

The second category, known as diffusion distillation, trains student models to distill the multi-step outputs of the original diffusion model into a single step. Progressive Distillation\cite{salimans2022progressive}, for instance, introduces binary time distillation, where the student model learns to predict the two-step output of the teacher model and subsequently assumes the teacher role in later iterations. Additionally, methods like Consistency Model\cite{song2023consistencymodels} and TRACT\cite{berthelot2023tract} incorporate self-consistency by using the student model, equipped with exponentially moving averaged weights, as a self-teacher to iteratively refine its predictions.
\subsection{Diffusion Bridge Models}
Diffusion bridge models have emerged as a compelling alternative to conditional diffusion models\cite{saharia2022image,dhariwal2021diffusion,saharia2022palette} for conditional image generation. Despite their diverse origins, models such as InDI\cite{delbracio2023inversion}, I$^2$SB\cite{liu20232}, and DDBM\cite{zhou2023denoising} can be unified under a shared framework\cite{chung2024direct,he2024consistency}. These models employ Doob's h-transform to adjust the forward process, ensuring it terminates at corrupted images. This adjustment allows the reverse process to initialize from the structurally informative corrupted images instead of Gaussian noise.

SDE-based samplers are prevalent in existing diffusion bridge models, including InDI, I$^2$SB, CDDB\cite{chung2024direct}, and IR-SDE\cite{luo2023image}. To accelerate the generative process, several PF-ODE-based techniques adapted from standard diffusion models have been applied. For instance, DDBM proposed a hybrid approach alternating between SDE and ODE samplers, utilizing the second-order Heun solver for the ODE step. I$^3$SB\cite{wang2024implicit} and DBIM\cite{zheng2024diffusion} adopted strategies from DDIM, transitioning the generative process from a Markovian to a non-Markovian framework. Additionally, Consistency Diffusion Bridge Model (CDBM)\cite{he2024consistency} applied consistency distillation from Consistency Models, achieving competitive results with only two generative steps.

Despite these advancements, the application of high-order ODE solvers to diffusion bridge models remains largely unexplored. To bridge this gap, we propose a novel method that integrates posterior sampling as a stochastic start and leverages the second-order Heun solver to solve the PF-ODE in diffusion bridge models. This approach demonstrates notable effectiveness and efficiency in image restoration and translation tasks.

\section{Method}
\subsection{Preliminaries on Diffusion Bridge Model}

Diffusion models establish a mapping between the data distribution $q_{\text{data}}$ and Gaussian noise by defining a continuous diffusion process $X_t\in \mathbb{R}^d$ indexed by a continuous time variable $t \in [0,T]$. This process can be modeled as the solution to the following forward SDE\cite{song2020score}:
\begin{equation}
\text{d}X_t=f\left(t\right)X_t\text{d}t+g\left(t\right)\text{d}w,
\label{eq:diffusion_sde}
\end{equation}
where $X_0 \sim q_{\text{data}}\left(X_0\right)$, $f\left(t\right)X_t$ represents the linear drift term, $g\left(t\right)$ controls the diffusion rate and $w$ is the Wiener process.

Diffusion bridge models are designed for image restoration and translation tasks, where a corrupted image $y$ is provided, and the goal is to sample from the conditional data distribution $q_{\text{data}}\left(X_0|y\right)$. CDBM\cite{he2024consistency} offers a unified framework that encompasses prominent diffusion bridge models, such as DDBM and I$^2$SB, despite originating from different perspectives. These models employ Doob's h-transform to adjust the dynamics of the diffusion process, resulting in the following forward SDE:
\begin{multline}
\text{d}X_t\\=\left[f\left(t\right)X_t+g^2\left(t\right)\nabla_{X_t}\log p_{T|t}\left(y|X_t\right)\right]\text{d}t+g\left(t\right)\text{d}w,
\label{eq:forward_sde}
\end{multline}
where $X_0 \sim q_{\text{data}}\left(X_0|y\right)$, and $p_{T|t}$ denotes the transition kernel from time $t$ to $T$ as defined by the original SDE (\ref{eq:diffusion_sde}). The term $\nabla_{X_t}\log p_{T|t}$ is expressed as:
\begin{equation}
\nabla_{X_t}\log p_{T|t}\left(y|X_t\right)=\frac{\left(\alpha_t/\alpha_T\right)y-X_t}{\alpha_t^2\left(\rho_T^2-\rho_t^2\right)},
\label{eq:ptT}
\end{equation}
with parameters defined as:
\begin{equation}
\alpha_t=\exp{\left(\int_0^tf\left(\tau \right)\text{d}\tau\right)},\quad\rho_t^2=\int_0^t\frac{g^2\left(\tau\right)}{\alpha_\tau^2}\text{d}\tau.
\end{equation}
We use  
$q_{t|0,y}\left(X_t|X_0,y\right)$ to denote the transition kernel from time $0$ to $t$, and $q_{t|y}\left(X_t|y\right)$ to represent the marginal distribution of $X_t$, both determined by the forward SDE (\ref{eq:forward_sde}) and conditioned on the corrupted image $y$. The forward SDE (\ref{eq:forward_sde})  ensures that the diffusion process almost surely converges to the fixed endpoint $y$, i.e., 
\begin{equation}
q_{T|y}\left(X_T|y\right)=\delta\left(X_T-y\right),
\label{eq:q_T}
\end{equation}
where $\delta$ represents the Dirac function. Additionally, the transition kernel $q_{t|0,y}\left(X_t|X_0,y\right)$ has a closed form expression, allowing efficient sampling in forward process:
\begin{equation}
q_{t|0,y}\left(X_t|X_0,y\right)=\mathcal{N}\left(X_t;a_ty+b_tX_0,c_t^2I\right),
\label{eq:qt|0y}
\end{equation}
where the parameters are given by:
\begin{subequations}
\begin{equation}
a_t=\frac{\rho_t^2\alpha_t}{\rho_T^2\alpha_T},
\end{equation}
\begin{equation}
b_t=\alpha_t\left(1-\frac{\rho_t^2}{\rho_T^2}\right),
\end{equation}
\begin{equation}
c_t^2=\alpha_t^2\rho_t^2\left(1-\frac{\rho_t^2}{\rho_T^2}\right).
\end{equation}
\end{subequations}

Samples from $q_{\text{data}}\left(X_0|y\right)$ can be generated by initializing $X_T=y$ and simulating the reverse-time process described by the reverse-time SDE\cite{zhou2023denoising}:
\begin{multline}
\text{d}X_t=f\left(t\right)X_t\text{d}t-\\g^2\left(t\right)\left(\nabla_{X_t}\log q_{t|y}\left(X_t|y\right)-\nabla_{X_t}\log p_{T|t}\left(y|X_t\right)\right)\text{d}t\\+g\left(t\right)\text{d}\overline{w},
\label{eq:reverse_sde}
\end{multline}
where $\overline{w}$ represents the reverse-time Wiener process. Both the forward SDE (\ref{eq:forward_sde}) and the reverse SDE (\ref{eq:reverse_sde}) share a common probability flow captured by the PF-ODE\cite{zhou2023denoising}:
\begin{multline}
\text{d}X_t=f\left(t\right)X_t\text{d}t-\\g^2\left(t\right)\left(\frac{1}{2}\nabla_{X_t}\log q_{t|y}\left(X_t|y\right)-\nabla_{X_t}\log p_{T|t}\left(y|X_t\right)\right)\text{d}t.
\label{eq:pf_ode}
\end{multline}
Direct regression to the score function $\nabla_{X_t}\log q_{t|y}\left(X_t|y\right)$ is impractical due to its singular behavior at time $t=T$. Instead, a data predictor $D_{\theta}\left(X_t,y,t\right)$ is used and trained by minimizing the loss:
\begin{equation}
\theta^*=\arg\min_{\theta}\mathbb{E}_{y,X_0,t,X_t}\omega\left(t\right)\Vert D_{\theta}\left(X_t,y,t\right)-X_0\Vert_2^2,
\end{equation}
where $\omega\left(t\right)$ is a positive weighting function and $X_t$ is sampled from $q_{t|0,y}\left(X_t|X_0,y\right)$. The score estimator $s_{\theta^*}\left(X_t,y,t\right)$ is derived from the trained data predictor $D_{\theta^*}\left(X_t,y,t\right)$ by
\begin{equation}
s_{\theta^*}\left(X_t,y,t\right)=\frac{1}{c_t^2}\left(b_tD_{\theta^*}\left(X_t,y,t\right)-X_t+a_ty\right),
\label{eq:score}
\end{equation}
and approximates the true score function $\nabla_{X_t}\log q_{t|y}\left(X_t|y\right)$ in the reverse process, following the principles of denoising bridge score matching\cite{zhou2023denoising}.

\begin{figure*}[!t]
\centering
\includegraphics[width=\textwidth]{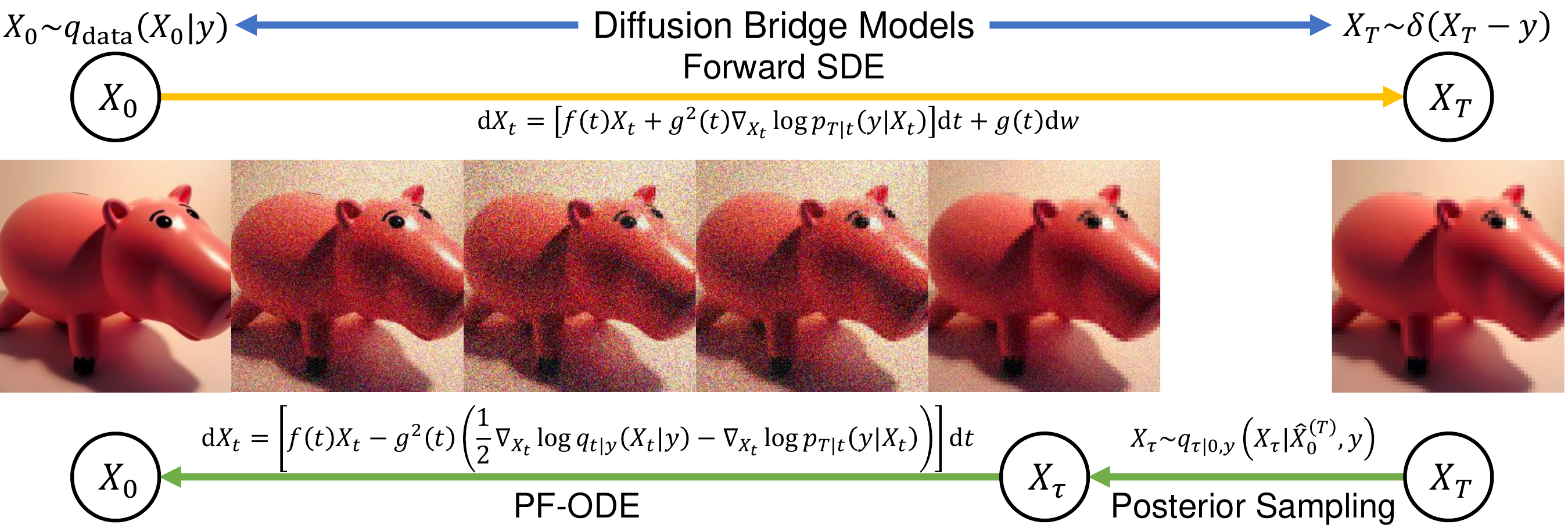}
\caption{Overview of the proposed ODE sampler with a stochastic start for diffusion bridge models. The forward SDE maps the conditional data distribution $q_{\text{data}}\left(X_0|y\right)$ to the Dirac distribution centered at the corrupted image $y$. In the reverse process, $X_T$ is initialized as $y$, posterior sampling is used to transition from time $T$ to $\tau$, and Heun's second-order solver is applied to solve the PF-ODE from time $\tau$ to 0.}
\label{fig:method}
\end{figure*}
\subsection{ODE Sampler with Stochastic Start}
High-order ODE samplers cannot be directly applied at the start of the reverse process in diffusion bridge models. While the reverse SDE (\ref{eq:reverse_sde}) remains well-defined at time $T$, the PF-ODE (\ref{eq:pf_ode}) exhibits singular behavior at $T$ due to the singularity in its non-linear drift term, as detailed in Theorem \ref{theorem:1} and \ref{theorem:2}. Detailed proofs are provided in the Appendix.
\begin{theorem}
\label{theorem:1}
At $t=T$, the non-linear drift term in the reverse SDE (\ref{eq:reverse_sde}) is well defined. Specifically,
\begin{multline}
\lim_{t\rightarrow T}\left[\nabla_{X_t}\log q_{t|y}\left(X_t|y\right)-\nabla_{X_t}\log p_{T|t}\left(y|X_t\right)\right]\\=-\frac{1}{\alpha_T^2\rho_T^2}\left(y-\alpha_T\hat{X}_0^{\left(T\right)}\right),
\label{eq:limit_sde}
\end{multline}
where the expected mean $\hat{X}_0^{\left(T\right)}$ is defined as:
\begin{equation}
\hat{X}_0^{\left(T\right)}=\int X_0q_{\text{data}}\left(X_0|y\right)\text{d}X_0.
\label{eq:hat_X0T}
\end{equation}
\end{theorem}
\begin{theorem}
\label{theorem:2}
At $t=T$, the non-linear drift term in the PF-ODE (\ref{eq:pf_ode}) becomes singular. Specifically, $\lim_{t\rightarrow T}\left[\frac{1}{2}\nabla_{X_t}\log q_{t|y}\left(X_t|y\right)-\nabla_{X_t}\log p_{T|t}\left(y|X_t\right)\right]$ does not exist.
\end{theorem}

Based on Theorem \ref{theorem:2}, transitioning from time $T$ to $\tau$ ($\tau<T$), where the PF-ODE (\ref{eq:pf_ode}) becomes well defined, is essential for utilizing high order ODE solvers in diffusion bridge models. While the reverse SDE (\ref{eq:reverse_sde}) is well-defined at time $T$, discretizing it in a single Euler-Maruyama step for this transition can lead to significant discretization errors, particularly when the step size $T-\tau$ is large. To address this issue, we employ a posterior sampling approach, generating $X_\tau$ from the distribution $q^{\text{post}}\left(X_\tau|y\right)$ defined as:
\begin{equation}
q^{\text{post}}\left(X_\tau|y\right)= q_{\tau|0,y}\left(X_\tau|\hat{X}_0^{\left(T\right)},y\right),
\label{eq:x_tau}
\end{equation}
where the expected mean $\hat{X}_0^{\left(T\right)}$ is obtained from the trained data predictor $D_{\theta^*}\left(X_t,y,t\right)$ as $\hat{X}_0^{\left(T\right)}=D_{\theta^*}\left(X_T,y,T\right)$, with $X_T=y$. The distribution $q_{\tau|0,y}\left(X_\tau|\hat{X}_0^{\left(T\right)},y\right)$ is defined by equation (\ref{eq:qt|0y}), with $t$ substituted by $\tau$ and $X_0$ approximated by $\hat{X}_0^{\left(T\right)}$. Compared to the distribution $q^{\text{EM}}\left(X_\tau|y\right)$ obtained by directly discretizing reverse SDE (\ref{eq:reverse_sde}) in a single Euler-Maruyama step, $q^{\text{post}}\left(X_\tau|y\right)$  aligns more closely with the true distribution $q_{\tau|0,y}\left(X_\tau|X_0,y\right)$ in terms of KL-Divergence. This observation is formalized in Theorem \ref{theorem:3}, with proof provided in the Appendix.
\begin{theorem}
\label{theorem:3}
The posterior sampling  distribution $q^{\text{post}}\left(X_\tau|y\right)$  more closely approximates the true distribution $q_{\tau|0,y}\left(X_\tau|X_0,y\right)$ compared to $q^{\text{EM}}\left(X_\tau|y\right)$, as measured by KL-Divergence. Specifically, the following inequality holds:
\begin{multline}
\mathbb{E}_{X_0\sim q_{data}\left(X_0|y\right)}\left[D_{KL}\left(q^{\text{post}}\left(X_\tau|y\right)||q_{\tau|0,y}\left(X_\tau|X_0,y\right)\right)\right]\\\leq\mathbb{E}_{X_0\sim q_{data}\left(X_0|y\right)}\left[D_{KL}\left(q^{\text{EM}}\left(X_\tau|y\right)||q_{\tau|0,y}\left(X_\tau|X_0,y\right)\right)\right]
\label{eq:inequity}
\end{multline}
\end{theorem}

Building on Theorem \ref{theorem:1}, \ref{theorem:2} and \ref{theorem:3},  we propose an ODE sampler with a stochastic start for diffusion bridge models, as illustrated in Fig. \ref{fig:method}. The reverse process begins by initializing $X_T$ with the given corrupted image $y$, i.e., $X_T=y$. To transition from time $T$ to $\tau$, posterior sampling is employed to reduce discretization errors. Subsequently, Heun's second order solver\cite{ascher1998computer} is applied to solve the PF-ODE (\ref{eq:pf_ode}) from time $\tau$ to 0, allowing the proposed sampler to achieve high perceptual quality with reduced NFEs. The complete procedure is outlined in Algorithm 1.
\begin{algorithm*}[t]
\renewcommand{\algorithmicrequire}
   {\textbf{Input:}}
   \renewcommand{\algorithmicensure}
   {\textbf{Output:}}
    \caption{ODE Sampler with Stochastic Start}
    \label{alg1}
    \begin{algorithmic}
		\REQUIRE Trained data predictor $D_{\theta^*}$, corrupted image $y$ \\time schedule $0=t_0<\dots< t_{i-1}< t_i< t_{i+1}<\dots<t_{N-1}=\tau< t_N=T$\\
        \textbf{Initialize}: $X_{t_N}\leftarrow y$
	\FOR{$n=N$ to $1$} 
		\IF{$n == N$}
		\STATE $\hat{X}_0^{\left(t_N\right)}\leftarrow D_{\theta^*}\left(X_{t_N},y,t_N\right)$  \hfill\COMMENT{SDE sampling at $T$}\\
        Sample $X_{t_{N-1}}$ from $q_{t_{N-1}|0,y}\left(X_{t_{N-1}}|\hat{X}_0^{(t_N)}, y\right)$
            \ELSE
            \STATE Get $s_{\theta^*}\left(X_{t_n},y,t_n\right)$ using $D_{\theta^*}\left(X_{t_n},y,t_n\right)$ with equation (\ref{eq:score})
            \hfill\COMMENT{ODE sampling}
            \\
            $d_n\leftarrow f\left(t_n\right)X_{t_n}-g^2\left(t_n\right)\left(\frac{1}{2}s_{\theta^*}\left(X_{t_n},y,t_n\right)-\nabla_{X_{t_n}}\log p_{T|t_n}\left(y|X_{t_n}\right)\right)$\\
            $X_{t_{n-1}}\leftarrow X_{t_n}+\left(t_{n-1}-t_n\right)d_n$
            \IF{$n\neq1$}
            \STATE
            Get $s_{\theta^*}\left(X_{t_{n-1}},y,t_{n-1}\right)$ using $D_{\theta^*}\left(X_{t_{n-1}},y,t_{n-1}\right)$ with equation (\ref{eq:score})
            \hfill\COMMENT{Heun's second order}
            \\
            $d_n'\leftarrow f\left(t_{n-1}\right)X_{t_{n-1}}-g^2\left(t_{n-1}\right)\left(\frac{1}{2}s_{\theta^*}\left(X_{t_{n-1}},y,t_{n-1}\right)-\nabla_{X_{t_{n-1}}}\log p_{T|t_{n-1}}\left(y|X_{t_{n-1}}\right)\right)$\\
            $X_{t_{n-1}}\leftarrow X_{t_n}+\frac{1}{2}\left(t_{n-1}-t_n\right)\left(d_n+d_n'\right)$
            \ENDIF
		\ENDIF 
            \ENDFOR
        \ENSURE $X_0=X_{t_0}$
    \end{algorithmic}
\end{algorithm*}
\subsection{Implementation Details}

We validated our proposed sampler on both image restoration and translation tasks. For image restoration, we utilized the pretrained models from I$^2$SB\cite{liu20232} and adhered to its time schedule during the generative process. Experiments were conducted on two types of degradations: 4$\times$ super-resolution with bicubic interpolation (sr4x-bicubic) and JPEG restoration with a quality factor of 10 (JPEG-10). Evaluations were performed on 10,000 randomly selected images from the validation dataset of ImageNet 256$\times$256\cite{deng2009imagenet}.

For image translation, we employed the pretrained Variance Preserving (VP) diffusion bridge models from DDBM\cite{zhou2023denoising}, following its time schedule during the generative process. We tested our method on two translation tasks: Edges→Handbags\cite{isola2017image} at a resolution of 64$\times$64 and DIODE-Outdoor\cite{vasiljevic2019diode} at a resolution of 256$\times$256. Evaluations were conducted on the entire training set for both Edges→Handbags and DIODE-Outdoor tasks, consistent with previous works\cite{zhou2023denoising, zheng2024diffusion}.

The number of generative steps 
$N$ was set to 20 (NFE=38) for image restoration tasks and 15 (NFE=28) for image translation tasks.

\begin{table}[t]
\begin{center}
\caption{
Quantitative results and computation time (per image) of tested methods for the sr4x-bicubic task. \textbf{Bold}: best, \underline{under}: second best.}
\setlength{\tabcolsep}{4mm}{
\begin{tabular}{lccc}
\hline
Method      &Time (s) &FID $\downarrow$ &CA $\uparrow$\\
\hline
ADM\cite{dhariwal2021diffusion}    &4.91     & 13.906  & 0.6618         \\
DDNM\cite{wang2022zero}   &2.77     & 13.997  & 0.6548      \\
DDRM\cite{kawar2022denoising}   &0.56     & 19.700 & 0.6350              \\
$\Pi$GDM\cite{song2023pseudoinverse} & 9.84&4.382&\textbf{0.7209}\\
DPS\cite{chung2022diffusion} & 121.65&10.251&0.6208\\
I$^2$SB (NFE=100)\cite{liu20232}&2.79&\underline{4.128}&\underline{0.7062}\\
ODES3 (ours, NFE=38)&1.04&\textbf{3.746}&0.7027\\
\hline
\end{tabular}}

\label{tab1}
\end{center}
\end{table}

\begin{table}[t]
\caption{
Quantitative results and computation time (per image) of tested methods for the jpeg-10 task. \textbf{Bold}: best, \underline{under}: second best.}
\begin{center}
\setlength{\tabcolsep}{4mm}{
\begin{tabular}{lccc}
\hline
Method      &Time (s) &FID $\downarrow$&CA $\uparrow$\\
\hline
DDRM\cite{kawar2022jpeg}   &5.42    & 19.977 & 0.6227             \\
$\Pi$GDM\cite{song2023pseudoinverse} & 11.39&6.137&0.7023\\
I$^2$SB (NFE=100)\cite{liu20232}&3.31&\underline{3.871}&\textbf{0.7172}\\
ODES3 (ours, NFE=38)&1.24&\textbf{3.169}&\underline{0.7122}\\
\hline
\end{tabular}}

\label{tab2}
\end{center}
\end{table}

\begin{figure*}[!t]
\centering
\includegraphics[width=\textwidth]{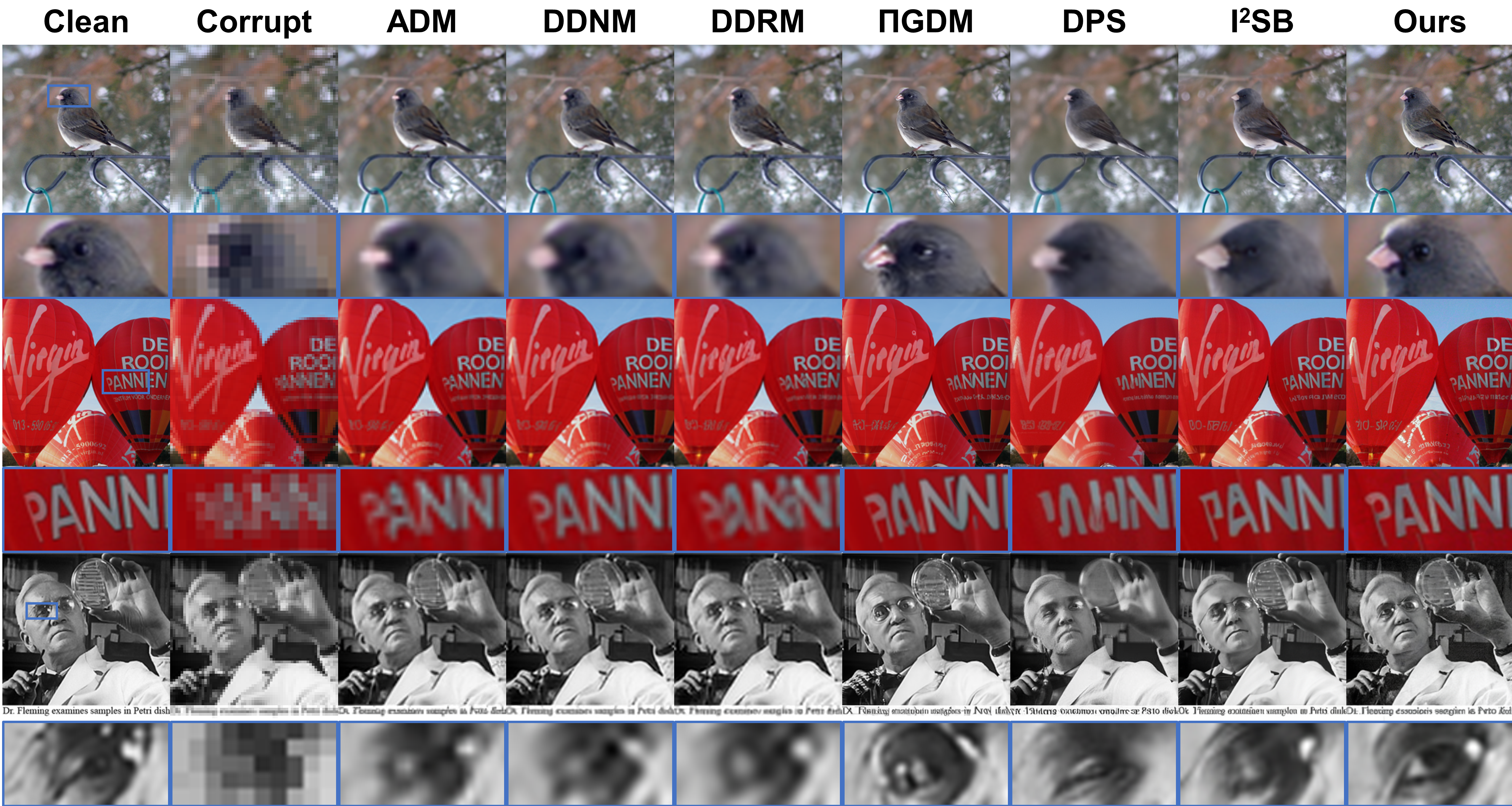}
\caption{Visualization results of tested methods for the sr4x-bicubic task. The details within the blue boxes are zoomed in for enhanced visual clarity. The NFE for I$^2$SB
is 100, and for our method is 38.}
\label{fig:sr4x_img}
\end{figure*}

\begin{figure*}[!t]
\centering
\includegraphics[width=\textwidth]{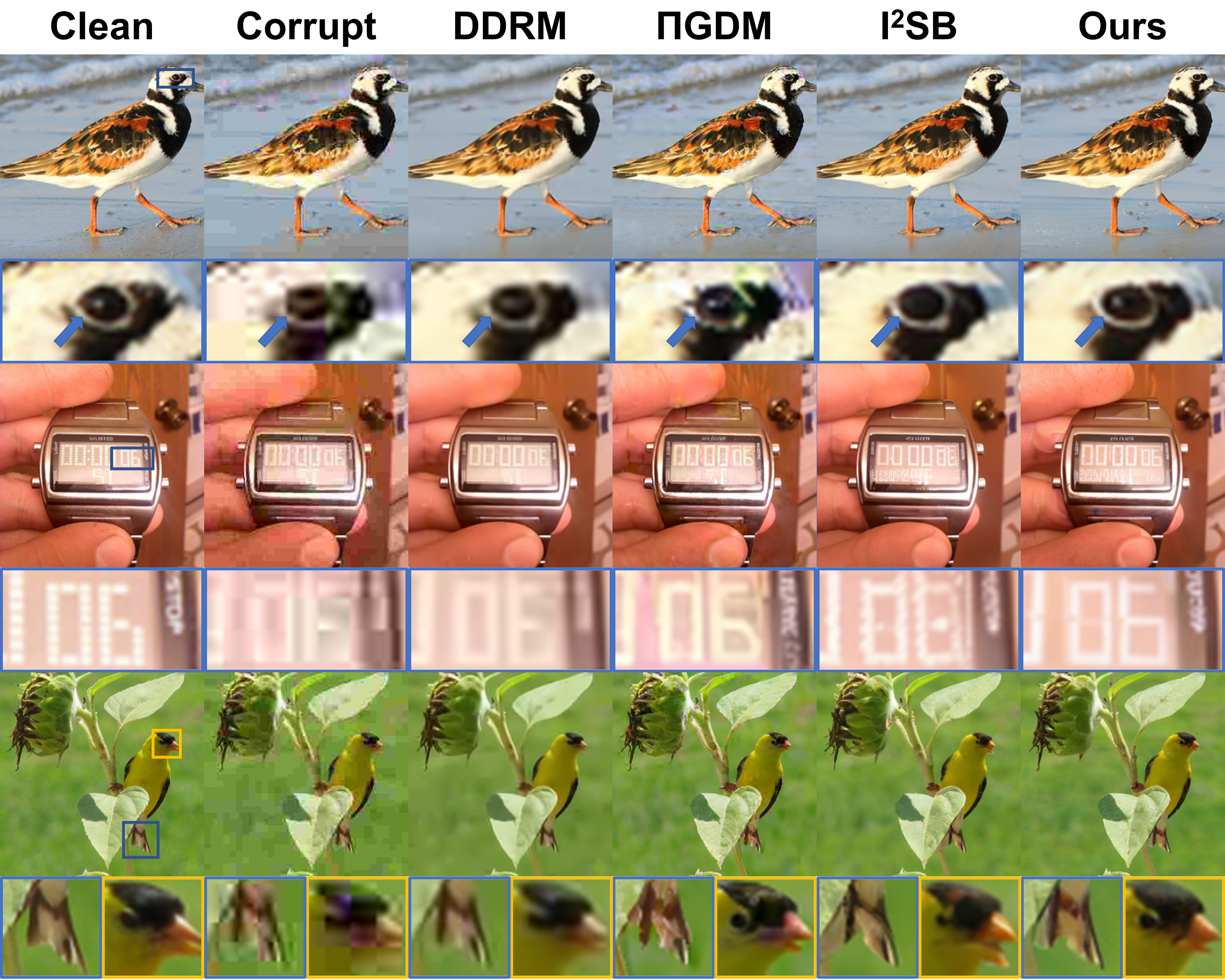}
\caption{Visualization results of tested methods for the JPEG-10 task. The details within the blue and yellow boxes are zoomed in for enhanced visual clarity. The NFE for I$^2$SB
is 100, and for our method is 38.}
\label{fig:jpeg_img}
\end{figure*}

\begin{table*}[t]
\caption{
Quantitative results in the image translation tasks. $^\dagger$Baseline results are taken directly from DDBM and DBIM, where they did not report the exact NFE. \textbf{Bold}: best, \underline{under}: second best.}
\begin{center}
\setlength{\tabcolsep}{4.5mm}{
\begin{tabular}{lccccccccc}
\hline
 &&\multicolumn{4}{c}{Edges$\rightarrow$ Handbags (64$\times $64)}&\multicolumn{4}{c}{DIODE-Outdoor (256$\times$256)} \\
\cmidrule(r){3-6}\cmidrule(r){7-10}
      &NFE&FID $\downarrow$&IS $\uparrow$        &LPIPS $\downarrow$&MSE $\downarrow$        &FID $\downarrow$&IS $\uparrow$        &LPIPS $\downarrow$&MSE $\downarrow$ \\
\hline
Pix2Pix\cite{isola2017image} & 1  & 74.8           & \textbf{4.24}        &0.356 & 0.209          & 82.4        &4.22 &0.556 &0.133    \\
DDIB\cite{su2022dual}& $\geq$40$^\dagger$       & 186.84       & 2.04        & 0.869  & 1.05          & 242.3        & 4.22&0.798&0.794  \\
SDEdit\cite{meng2021sdedit}& $\geq$40   & 26.5          &3.58          & 0.271&0.510            & 31.14          & 5.70          &  0.714 & 0.534    \\
Rectified Flow\cite{liu2022flow} & $\geq$40&25.3&2.80&0.241&0.088&77.18&5.87&0.534&0.157\\
I$^2$SB\cite{liu20232} & $\geq$40&7.43&3.40&0.244&0.191&9.34&5.77&0.373&0.145\\
DDBM (VP)\cite{zhou2023denoising} & 118 & 1.83&\underline{3.73}&0.142&0.040&4.43&\textbf{6.21}&0.244&0.084\\
DBIM\cite{zheng2024diffusion} & 100 &\underline{0.89}&3.62&\underline{0.100}&\underline{0.006}&\underline{2.57}&\underline{6.06}&\textbf{0.198}&\textbf{0.018}\\
ODES3 (ours)&28&\textbf{0.54}&3.65&\textbf{0.097}&\textbf{0.005}&\textbf{2.29}&5.92&\underline{0.203}&\textbf{0.018}\\
\hline

\end{tabular}}
\label{tab3}
\end{center}
\end{table*}

\begin{figure*}[!t]
\centering
\includegraphics[width=\textwidth]{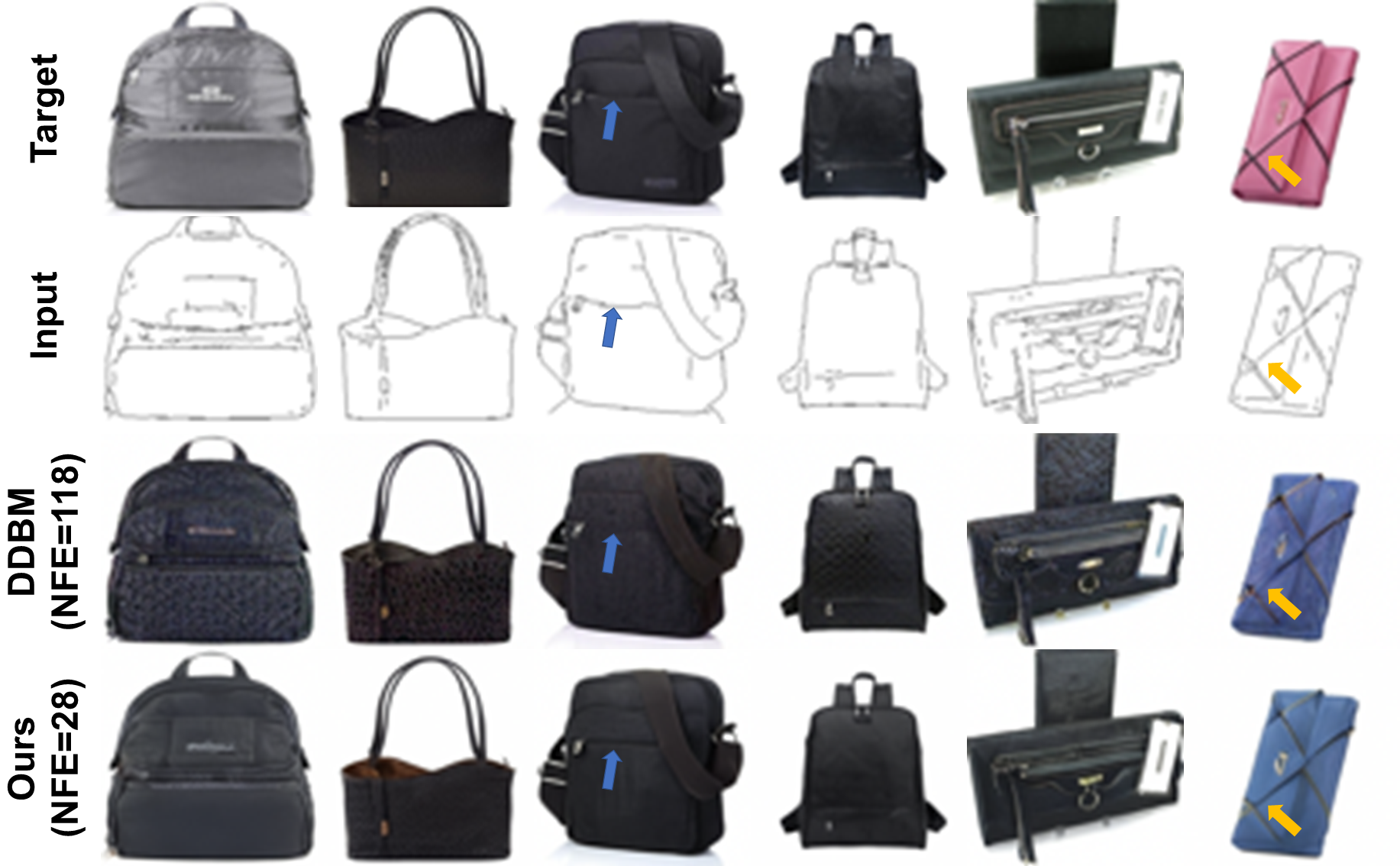}
\caption{Visualization results for the Edges$\rightarrow$Handbags (64$\times$64) task.}
\label{fig:e2h_img}
\end{figure*}

\begin{figure*}[!t]
\centering
\includegraphics[width=\textwidth]{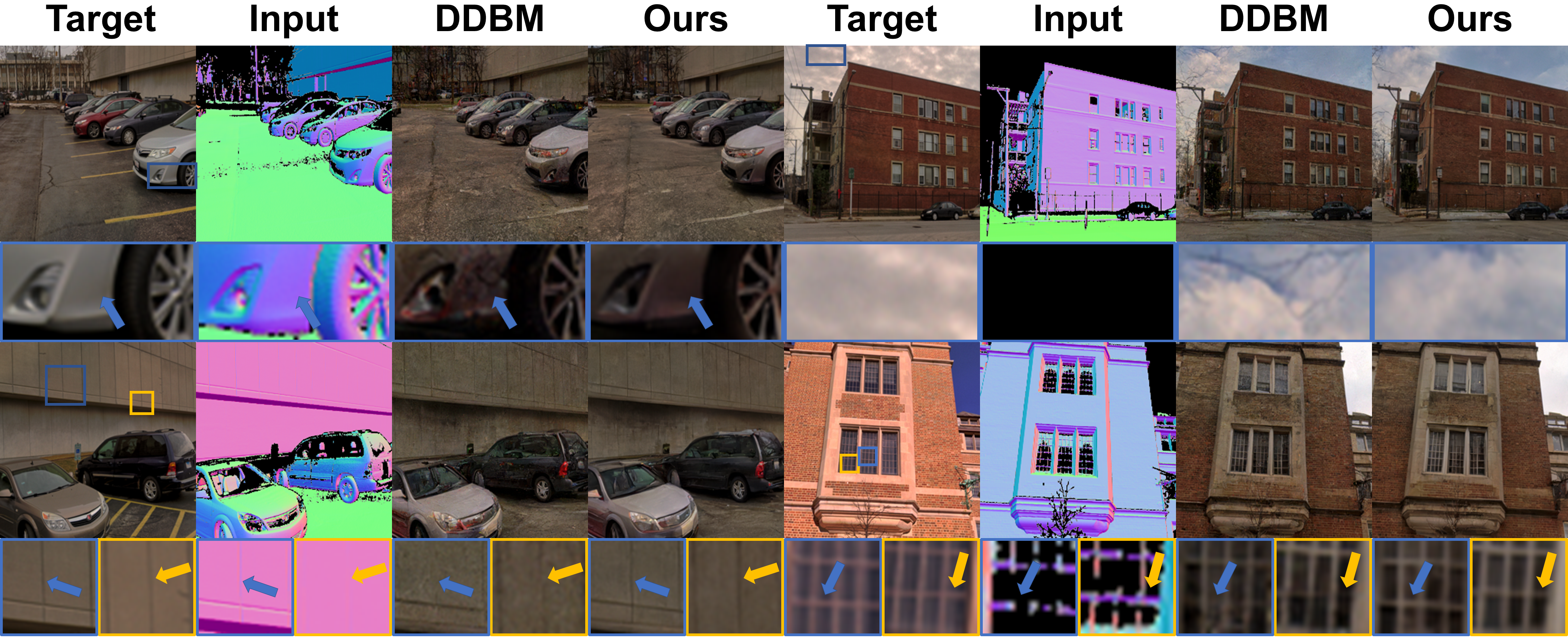}
\caption{Visualization results for the DIODE-Outdoor (256$\times$256) task. The details within
the blue and yellow boxes are zoomed in for enhanced visual clarity. The NFE for DDBM is
118, and for our method is 28.}
\label{fig:DIODE_img}
\end{figure*}
\section{Results}
\subsection {Image Restoration Tasks}
In image restoration tasks, we compared our proposed sampler with the SDE-based sampler from I$^2$SB\cite{liu20232}, using the same pretrained I$^2$SB models. We also included comparisons with the conditional diffusion model from ADM\cite{dhariwal2021diffusion}, and several diffusion-based plug-and-play models, including DDNM\cite{wang2022zero}, DDRM\cite{kawar2022denoising,kawar2022jpeg}, $\Pi$GDM\cite{song2023pseudoinverse}, and DPS\cite{chung2022diffusion}. For quantitative evaluation, we calculated FID\cite{heusel2017gans} to assess perceptual quality and included Classifier Accuracy (CA) using a pretrained ResNet-50\cite{he2016deep}, following previous work\cite{liu20232}. The results are presented in TABLE \ref{tab1} for the sr4x-bicubic task and in TABLE \ref{tab2} for the JPEG-10 task. Baseline values were computed using the official implementations of these methods with default hyperparameters. All experiments were conducted on a single A100 GPU, and the computation time for all tested methods is included in the tables. Representative results are visualized in Fig. \ref{fig:sr4x_img} for the sr4x-bicubic task and Fig. \ref{fig:jpeg_img} for the JPEG-10 task.

In the sr4x-bicubic task (TABLE \ref{tab1}), our method, ODES3, achieved the best FID among all tested approaches. Compared to the 100-step I$^2$SB, our sampler reduced FID by 9\% and offered a 2.7$\times$ acceleration in computation time while maintaining comparable CA. When compared to ADM, DDNM, DDRM, and DPS, our method demonstrated superior results, with a 6 to 16-point decrease in FID and a 0.04 to 0.08 increase in CA. Although $\Pi$GDM achieved the highest CA in this task, it required a known forward operator to incorporate data consistency during inference, exhibited a worse FID, and was 9.5× slower than our method. The superior performance of our sampler is further confirmed by the visualization results in Fig. \ref{fig:sr4x_img}, where our method shows enhanced detail restoration compared to all comparison methods, particularly in regions such as the eyes of birds and humans, as well as in text characters.

In the JPEG-10 task (TABLE \ref{tab2}), our method achieved the best FID among all tested approaches. Compared to the 100-step I$^2$SB, our sampler reduced FID by 18\% and achieved a 2.7× acceleration in computation time while maintaining comparable CA. When compared to DDRM and $\Pi$GDM, our method demonstrated superior performance, with a 3 to 17-point reduction in FID and a 0.01 to 0.09 improvement in CA. The exceptional performance of our sampler is further supported by the visualization results in Fig. \ref{fig:jpeg_img}, where our method exhibits enhanced detail restoration compared to all comparison methods, particularly in regions such as the eyes and tails of birds, as well as the numbers on the clock.
\subsection{Image Translation Tasks}

In image translation tasks, we compared our sampler with the hybrid high-order sampler in DDBM\cite{zhou2023denoising} and the non-Markovian sampler in DBIM\cite{zheng2024diffusion}, using the same pretrained DDBM (VP) models. Following previous work\cite{zhou2023denoising}, we also included comparisons with Pix2Pix\cite{isola2017image}, DDIB\cite{su2022dual}, SDEdit\cite{meng2021sdedit}, Rectified Flow\cite{liu2022flow}, and I$^2$SB\cite{liu20232}. We used FID\cite{heusel2017gans} to evaluate perceptual quality and also reported Inception Scores (IS)\cite{barratt2018note}, Learned Perceptual Image Patch Similarity (LPIPS)\cite{zhang2018perceptual}, and Mean Squared Error (MSE) for all tested methods. The results are summarized in TABLE \ref{tab3}, with baseline results directly taken from DDBM\cite{zhou2023denoising} and DBIM\cite{zheng2024diffusion}. Representative results for DDBM and our method are visualized in Fig. \ref{fig:e2h_img} for the Edges$\rightarrow$Handbags (64×64) task and in Fig. \ref{fig:DIODE_img} for the DIODE-Outdoor (256×256) task.

Our method achieved the best FID among all tested approaches in both image translation tasks. Compared to DDBM, our method exhibited slightly lower IS but delivered superior performance in terms of FID, LPIPS, and MSE. Notably, our sampler reduced FID by 70\% in the Edges$\rightarrow$Handbags task and 48\% in the DIODE-Outdoor task, with a 4.2$\times$ acceleration in NFE. Visualization results in Fig. \ref{fig:e2h_img} and \ref{fig:DIODE_img} reveal that DDBM tends to generate artificial artifacts, such as checkerboard patterns on bags and tree branches floating in the sky. In contrast, our method effectively mitigates these issues, achieving better detail restoration, particularly in regions such as bag zippers, gaps in walls, and window frames.

When compared to DBIM, our method achieved similar IS, LPIPS, and MSE  but reduced FID by 39\% in the Edges→Handbags task and 11\% in the DIODE-Outdoor task, with a 3.6$\times$ acceleration in NFE. Against Pix2Pix, our method achieved substantial FID reductions of 99\% in the Edges→Handbags task and 97\% in the DIODE-Outdoor task, although with a slightly lower IS in the Edges→Handbags task. Additionally, our approach consistently outperformed other methods across all four metrics in both tasks.

\section{Conclusion and Future Work}

In conclusion, we presented a high-order ODE sampler with a stochastic start for diffusion bridge models. Recognizing that the PF-ODE exhibits singular behavior at the start of the generative process, while the reverse SDE remains well-defined, we introduced a stochastic start and employed posterior sampling to mitigate discretization errors. Following this, we utilized Heun's second-order solver to solve the PF-ODE, enabling our sampler to achieve high perceptual quality with reduced NFEs.

Our sampler is fully compatible with pretrained diffusion bridge models, requiring no additional training, and was validated on both image restoration and translation tasks using pretrained models from I$^2$SB and DDBM. Compared to the original samplers used in I$^2$SB and DDBM, our sampler achieved better FID with fewer NFEs and provided superior detail restoration in visualizations. Additionally, our method demonstrated significant improvements over other state-of-the-art methods, including ADM, DDNM, DDRM, $\Pi$GDM, and DPS in image restoration tasks, and Pix2Pix, DDIB, SDEdit, Rectified Flow, I$^2$SB, and DBIM in image translation tasks.

This work focused on the starting strategy for the generative process of diffusion bridge models, leaving the exploration of alternative high-order ODE solvers as a future direction. While we employed Heun's second-order solver and demonstrated its strong performance, replacing it with other high-order ODE solvers, known for their effectiveness in diffusion models, holds potential for further improving the performance of our sampler. We plan to investigate these possibilities in future research.

\section*{Appendix}
\subsection{Proofs}
\subsubsection{Proof for Theorem \ref{theorem:1} and \ref{theorem:2}}
For any $t\in\left(0,T\right)$, we proceed the score function $\nabla_{X_t}\log q_{t|y}\left(X_t|y\right)$ as follows:
\begin{equation}\label{eq:proceed_score}
\begin{aligned}
&\nabla_{X_t}\log q_{t|y}\left(X_t|y\right)\\&=\frac{1}{q_{t|y}\left(X_t|y\right)}\nabla_{X_t}q_{t|y}\left(X_t|y\right),\\
&=\frac{1}{q_{t|y}\left(X_t|y\right)}\nabla_{X_t}\int q_{t|0,y}\left(X_t|X_0,y\right)q_{\text{data}}\left(X_0|y\right)\text{d}X_0,\\
&=\frac{1}{q_{t|y}\left(X_t|y\right)}\int \left(\nabla_{X_t}q_{t|0,y}\left(X_t|X_0,y\right)\right)q_{\text{data}}\left(X_0|y\right)\text{d}X_0,\\
&=\frac{\int \left(\frac{\left(a_ty+b_tX_0\right)-X_t}{c_t^2}q_{t|0,y}\left(X_t|X_0,y\right)\right)q_{\text{data}}\left(X_0|y\right)\text{d}X_0}{q_{t|y}\left(X_t|y\right)},\\
&=\frac{a_ty+b_t\int X_0\frac{q_{\text{data}}\left(X_0|y\right)q_{t|0,y}\left(X_t|X_0,y\right)}{q_{t|y}\left(X_t|y\right)}\text{d}X_0-X_t}{c_t^2},\\
&=-\frac{1}{c_t^2}\left(X_t-\left(a_ty+b_t\hat{X}_0^{\left(t\right)}\right)\right),
\end{aligned}
\end{equation}
where the expected mean $\hat{X}_0^{(t)}$ is defined as: 
\begin{equation}
\hat{X}_0^{(t)}=\int X_0q_{0|t,y}\left(X_0|X_t,y\right)\text{d}X_0,
\end{equation}
and $q_{0|t,y}\left(X_0|X_t,y\right)$ is given by Bayesian rule:
\begin{equation}
q_{0|t,y}\left(X_0|X_t,y\right)=\frac{q_{data}\left(X_0|y\right)q_{t|0,y}\left(X_t|X_0,y\right)}{q_{t|y}\left(X_t|y\right)}.
\end{equation}. 

Using the expression for $\nabla_{X_t}\log p_{T|t}$ in equation (\ref{eq:ptT}), the non-linear drift term in the reverse SDE (\ref{eq:reverse_sde}) is expressed as:
\begin{multline}
\nabla_{X_t}\log q_{t|y}\left(X_t|y\right)-\nabla_{X_t}\log p_{T|t}\left(y|X_t\right)\\=-\frac{1}{\alpha_t^2\rho_t^2}\left(X_t-\alpha_t\hat{X}_0^{\left(t\right)}\right).
\end{multline}
Letting $t\rightarrow T$ with $X_T=y$, we establish equation (\ref{eq:limit_sde}). Furthermore, the non-linear drift term in PF-ODE (\ref{eq:pf_ode}) can be expressed in terms of the corresponding term in the reverse SDE (\ref{eq:reverse_sde}) and the score function as follow:
\begin{multline}
\frac{1}{2}\nabla_{X_t}\log q_{t|y}\left(X_t|y\right)-\nabla_{X_t}\log p_{T|t}\left(y|X_t\right)\\=-\frac{X_t-\alpha_t\hat{X}_0^{\left(t\right)}}{\alpha_t^2\rho_t^2}-\frac{1}{2}\nabla_{X_t}\log q_{t|y}\left(X_t|y\right).
\end{multline}
As $t\rightarrow T$, the term $\frac{1}{2}\nabla_{X_t}\log q_{t|y}\left(X_t|y\right)-\nabla_{X_t}\log p_{T|t}\left(y|X_t\right)$ does not converge, due to the singularity of the score function at time $T$.
\subsubsection{Proof for Theorem
\ref{theorem:3}}
To proof Theorem \ref{theorem:3}, we start by formulating the following optimization problem:
\begin{subequations}
\begin{multline}
q^*\left(X_\tau|y\right)=\arg\min_{q\left(X_\tau|y\right)}\mathbb{E}_{X_0\sim q_{\text{data}}\left(X_0|y\right)}\\\left[\text{D}_\text{KL}\left(q\left(X_\tau|y\right)||q_{\tau|0,y}\left(X_\tau|X_0,y\right)\right)\right],
\end{multline}
\begin{equation}
\text{s.t.}\quad q\left(X_\tau|y\right)=\mathcal{N}\left(X_\tau|\mu_\tau\left(y\right),\sigma_\tau^2\left(y\right)I\right),
\end{equation}
\label{eq:opt}
\end{subequations}
where $q_{\tau|0,y}\left(X_\tau|X_0,y\right)$ is defined in equation (\ref{eq:qt|0y}) with $t$ substituted by $\tau$, and $\mu_\tau\left(y\right)$ and $\sigma_\tau\left(y\right)$ are the parameters to be optimized. Since both $q\left(X_\tau|y\right)$ and $q_{\tau|0,y}\left(X_\tau|X_0,y\right)$ are Gaussian distributions, the KL-divergence can be explicitly computed. The objective function becomes:
\begin{multline}
\mathbb{E}_{X_0\sim q_{\text{data}}\left(X_0|y\right)}\text{D}_\text{KL}\left(q\left(X_\tau|y\right)||q_{\tau|0,y}\left(X_\tau|X_0,y\right)\right)\\
=\frac{d}{2}\left(\frac{\sigma_\tau^2\left(y\right)}{c_\tau^2}-1-\ln \frac{\sigma_\tau^2\left(y\right)}{c_\tau^2}\right)\\+\frac{1}{2c_\tau^2}\mathbb{E}_{X_0\sim q_{\text{data}}\left(X_0|y\right)}\Vert a_\tau y+b_\tau X_0-\mu_\tau\left(y\right)\Vert_2^2,
\end{multline}
where $d$ denotes the data dimension. The optimal parameters $\mu_\tau^*\left(y\right)$ and $\sigma_\tau^*\left(y\right)$ that minimize the objective function are:
\begin{subequations}
\begin{equation}
\mu_\tau^*\left(y\right)=a_\tau y+b_\tau\hat{X}_0^{\left(T\right)},
\end{equation}
\begin{equation}
\sigma_\tau^*\left(y\right)=c_\tau,
\end{equation}
\end{subequations}
where the expected mean $\hat{X}_0^{\left(T\right)}$ is defined in equation (\ref{eq:hat_X0T}). Since $\mu_\tau^*\left(y\right)$ and $\sigma_\tau^{*^2}\left(y\right)I$ correspond to the mean and covariance matrix of the Gaussian distribution $q^{\text{post}}\left(X_\tau|y\right)$, we conclude:
\begin{equation}
q^*\left(X_\tau|y\right)=q^{\text{post}}\left(X_\tau|y\right),
\label{eq:q_star}
\end{equation}
demonstrating that $q^{\text{post}}\left(X_\tau|y\right)$ is the optimal Gaussian approximation to the true distribution $q_{\tau|0,y}\left(X_\tau|X_0,y\right)$ in terms of KL-divergence.

Next, discretizing the reverse SDE (\ref{eq:reverse_sde}) from time $T$ to $\tau$ using a single Euler-Maruyama step results in:
\begin{multline}
X_\tau-y=\left(f\left(T\right)y+g^2\left(T\right)\frac{y-\alpha_T\hat{X}_0^{\left(T\right)}}{\alpha_T^2\rho_T^2}\right)\left(\tau-T\right)\\+g\left(T\right)\sqrt{T-\tau}\epsilon,
\end{multline}
where $\epsilon\sim\mathcal{N}\left(0,I\right)$. Thus, $q^{\text{EM}}\left(X_\tau|y\right)$ is also a Gaussian distribution. By combining this result with equations (\ref{eq:opt}) and (\ref{eq:q_star}), we conclude that the inequity (\ref{eq:inequity}) is satisfied, completing the proof of Theorem \ref{theorem:3}.

\vfill

\end{document}